\documentclass{article}

\usepackage[utf8]{inputenc}
\usepackage[LFE,LAE,T1]{fontenc}
\usepackage{arabtex}
\usepackage{utf8}
\setcode{utf8}

\usepackage{placeins}
\usepackage{subcaption}
\usepackage{arxiv}

\usepackage[utf8]{inputenc} 
\usepackage[T1]{fontenc}    
\usepackage{hyperref}       
\usepackage{url}            
\usepackage{booktabs}       
\usepackage{amsfonts}       
\usepackage{nicefrac}       
\usepackage{microtype}      
\usepackage{lipsum}		
\usepackage{graphicx}
\usepackage{doi}
\usepackage{CJKutf8}
\usepackage{multirow}

\title{A Deep CNN Architecture with Novel Pooling Layer Applied to Two Sudanese Arabic Sentiment Datasets}

\author{\mbox{\href{https://orcid.org/0000-0002-3106-669X}{\includegraphics[scale=0.06]{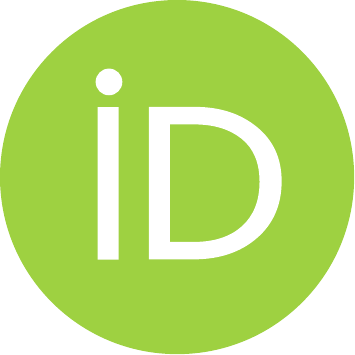}\strut}}\hspace{1mm}Mustafa Mhamed \\
	School of Information Science and Technology\\
	Northwest University\\
	Xi’an 710127, China \\
	\texttt{mustafamhamed@stumail.nwu.edu.cn} \\
	\And
	\mbox{\href{https://orcid.org/0000-0002-5549-5691}{\includegraphics[scale=0.06]{orcid.pdf}\strut}}\hspace{1mm}Richard Sutcliffe\thanks{Corresponding author} \\
	School of Information Science and Technology\\
	Northwest University\\
	Xi’an 710127, China \\
	School of Computer Science and Electronic Engineering \\ University of Essex \\
	Wivenhoe Park, Colchester CO4 3SQ, UK\\
	\texttt{rsutcl@nwu.edu.cn, rsutcl@essex.ac.uk} \\
	\And
	\mbox{\href{https://orcid.org/0000-0003-0572-641X}{\includegraphics[scale=0.06]{orcid.pdf}\strut}}\hspace{1mm}Xia Sun\thanks{Corresponding author} \\
	School of Information Science and Technology\\
	Northwest University\\
	Xi’an 710127, China \\
	\texttt{raindy@nwu.edu.cn} \\
	\And
	\mbox{\href{https://orcid.org/0000-0002-0706-2103}{\includegraphics[scale=0.06]{orcid.pdf}\strut}}\hspace{1mm}Jun Feng\\
	School of Information Science and Technology \\
	Northwest University\\
	Xi’an 710127, China \\
	\texttt{fengjun@nwu.edu.cn} \\
\And
	\mbox{\href{https://orcid.org/0000-0002-7182-9639}{\includegraphics[scale=0.06]{orcid.pdf}\strut}}\hspace{1mm}Eiad Almekhlafi \\
	School of Information Science and Technology\\
	Northwest University\\
	Xi’an 710127, China \\
	\texttt{ealmekhlafi@stumail.nwu.edu.cn} \\
	\And
	\mbox{\href{https://orcid.org/0000-0003-0906-9605}{\includegraphics[scale=0.06]{orcid.pdf}\strut}}\hspace{1mm}Ephrem Afele Retta \\
	School of Information Science and Technology\\
	Northwest University\\
	Xi’an 710127, China \\
	\texttt{afele@stumail.nwu.edu.cn}
}

\date{}

\renewcommand{\headeright}{mustafa mhamed et al.}
\renewcommand{\undertitle}{}
\renewcommand{\shorttitle}{Sudanese
Arabic Sentiment Datasets}

\hypersetup{
pdftitle={A Deep CNN Architecture with Novel Pooling Layer Applied to Two Sudanese Arabic Sentiment Datasets},
pdfsubject={cs.CL, cs.LG},
pdfauthor={Mustafa Mhamed, Richard Sutcliffe, Xia Sun, Jun Feng, Eiad Almekhlafi, Ephrem Afele Retta},
pdfkeywords={Arabic Sentiment Analysis, Natural Language Processing, NLP, SudSenti2, SudSenti3, Sentiment Classification Model, Mean Max Average Pooling}
}
\begin{document}
\maketitle

\begin{abstract}
Arabic sentiment analysis has become an important research field in recent years. Initially, work focused on Modern Standard Arabic (MSA), which is the most widely-used form. Since then, work has been carried out on several different dialects, including Egyptian, Levantine and Moroccan. Moreover, a number of datasets have been created to support such work. However, up until now, less work has been carried out on Sudanese Arabic, a dialect which has 32 million speakers. In this paper, two new publicly available datasets are introduced, the 2-Class Sudanese Sentiment Dataset (SudSenti2) and the 3-Class Sudanese Sentiment Dataset (SudSenti3). Furthermore, a CNN architecture, SCM, is proposed, comprising five CNN layers together with a novel pooling layer, MMA, to extract the best features. This SCM+MMA model is applied to SudSenti2 and SudSenti3 with accuracies of 92.75\% and 84.39\%. Next, the model is compared to other deep learning classifiers and shown to be superior on these new datasets. Finally, the proposed model is applied to the existing Saudi Sentiment Dataset and to the MSA Hotel Arabic Review Dataset with accuracies 85.55\% and 90.01\%.
\end{abstract}

 
\section{Introduction}

Sentiment analysis is an important field because it enables us to discover voices and opinions relating to topics of interest in a particular context, for example, views about political issues in elections, or opinions about products or ways of providing services.
With the emergence of participatory web services in areas such as education and health, there is a need for sentiment analysis in identifying problems and hence upgrading quality standards.
Recently, the spread of Arabic content, especially on social media, and the application of artificial intelligence and deep learning in analyzing Arabic sentiments, has led researchers to delve deeper into Arabic text. Initially, this work has been carried out on Modern Standard Arabic (MSA). However, more recent work has also been concerned with the regional Arabic dialects that are often used in everyday informal communications.

\begin{table}[ht]
        \centering
        \caption{Dialects of Arabic (derived from istizada.com).}
         \begin{tabular}{ccc}
		\hline
         Dialect & Areas Spoken & Number of Speakers \\\hline 
         Gulf & Bahrain, Kuwait, Oman, Qatar, Saudi Arabia, UAE & 36,056,000\\ \hline
          \textbf{Sudanese} & \textbf{Sudan, Southern Egypt} & \textbf{31,940,300 } \\\hline
		
         Hassaniya	&Mauritania, southern Morocco, south western Algeria, Western Sahara &8,842,800\\\hline
						
        Levantine & Lebanon, Jordan, Palestine, Syria &36,188,500 \\\hline
        Maghrebi&Algeria, Libya, Morocco, Tunisia &32,608,700 \\\hline
		
        Mesopotamian / Iraqi & Iraq, eastern Syria & 15,655,900 \\\hline
	
        Yemeni	&Yemen, Somalia, Djibouti, southern Saudi Arabia  &14,360,000  \\		
		\hline 
           		\hline 
           	\end{tabular}
           	\label{tb1}
           \end{table}   
           
          \begin{table*}
	\caption{Dialect Variation (based on \cite{oueslati2020review} with Sudanese additions).}\label{tb2} 
\centering	
 \begin{tabular}{lllll}
		\toprule
			MSA word & Dialectical word & Arabizi &Country & English Equivalent
		   \\
		\midrule
		
 \RL{حلو , جميل }\\Jameel & \RL{حلو}&7elew & Lebanon &nice  \\
 & \RL{حلو }&7ilew & Saudi &  \\

& \RL{حلو }&7low/hlow & Tunisia &  \\

& \RL{سمح}& Samh & Sudan &   \\\hline

\RL{جدآ }\\ Jiddan & \RL{ كتير }& ktir & Lebanon & very wide  \\

& \RL{ وايد}& wayed & Emirate & \\

& \RL{ أوي }&2awi & Egypt &  \\

& \RL{ كيتير }&kityer & Sudan &  \\

& \RL{ برشا }& barcha & Tunisia & \\\hline

\RL{دراجة }\\ Darraja & \RL{ بسكلات}& besklet & Tunisia & bicycle  \\

& \RL{ درافة }& darraga & Egypt & \\

& \RL{ عجلة}& Ajala & Sudan & \\

 		\bottomrule
	\end{tabular}
\end{table*}

\begin{table*}
	\caption{Examples of differences between MSA and the Sudanese dialect.}\label{tb3}
	\centering
\resizebox{\columnwidth}{!}{%
\begin{tabular}{lrr}	
		\toprule
	English language  & Standard Arabic & Sudanese dialect \\

	\toprule
Be careful today don't go out of the room&
\RL { كن حذراً اليوم لا تخرج من الغرفة } &
\RL { اعمل حسابك الليلة ما تطلع من الأوضه} \\
Keep going down this road until you find a pharmacy&
\RL {  إستمر في السير في هذا الطريق حتي تجد صيدلية } &
\RL{  أمشي دوغري لغاية تشوف اجزخانة }\\
Leave drinking too much coffee not good for your health&\RL { اترك شرب القهوة الـكثيرة غير مفيدة لصحتك } & \RL { سيب الجبنة الـكتيرة ما كويسة لجسمك } \\
		\bottomrule
	\end{tabular}
	}
\end{table*}

\begin{table*}
 \caption{Previous work on sentiment analysis for different Arabic dialects.}\label{tb4}
\centering
 	\begin{tabular}{@{}lcccc@{} } 
 		\toprule
 	  Paper   &   Arabic dialect &  Split  & Model & Result  \\
 	\midrule
 	\cite{alwehaibi2018comparison}  & Saudi (2C)& 80+20 & LSTM-RNN &
 		  93.5\% \\\hline
 		  \cite{alahmary2021semiautomatic}  & Saudi (3C)& 90+10 & CNN &
 		  86.54\% \\\hline
 		\cite{tabii2018big}  & Moroccan (2C) & 90+10
 		   & Majority Voting & 83.45\% \\\hline
 		\cite{lulu2018automatic}  & Egyptian, Iraqi and Levantine (3C) & 80+10+10
 		   & LSTM & 71.4 \% \\\hline
 	\cite{al2018sentiment} & Jordanian (2C) & 90+10
 	 & Ensemble & 93.4\% \\
 	 \hline
 	 \cite{al2019sentiment} & Lebanon (2C) & 80+20 & LR & 89.80\% \\
 		\hline
 		\cite{abdelli2019sentiment} &  Algerian (2C) & 85+15 & SVM & 0.86\%\\
 		
 		 	\hline
 	\cite{mulki2019empirical} & Tunisian (2C) & 80+10+10 & Deep-LSTM & 90.00\% \\
  \hline
 		\cite{mulki2019empirical} & JEG, TAC and TSAC (2C) & 90+10 & Tw-StAR & 82.08\%  \\
 		\hline
 		
 		\cite{mhamed2021improving} & Egyptian, MSA (2C)(n-C) & 80+10+10 & MC1, MC2 & 92.96\%  \\
 		\hline
 	
 		\cite{al2020meter} & Modern Standard Arabic & 85+15 & BiGRU & 94.32\% \\
 		\hline
       \cite{addi2020sampling} & Modern Standard Arabic & 80+20 & RF+SMOTE & 96.00\% \\
 						
 		\bottomrule
 	\end{tabular}

 \end{table*}

There are a significant number of variations between dialects and MSA in terms of language:
\begin{itemize}

\item MSA has a dual form of short vowel, omitted in written text, in addition to the singular and plural vowel forms, for masculine and feminine. Dialects often do not create such uniqueness between the sexes; instead they have an open system which is more complex than MSA, allowing the prefix and suffix to be attached to a base, and pronouns to function as indirect objects.
\item MSA has a complex system of grammar, which dialects sometimes lack. Dialects may not use diacritics, whereas most instances in MSA are represented with directly written diacritics since adverbs and objects are expressed using suffixes.

\item Arabic vocabulary varies according to the dialect used. For example, `money' in MSA is \RL{مال} , whereas in dialects it is  \RL{نقود , فلوس , مصاري , قروش}.

Similarly, `beautiful place' is \RL{المكان جميل} (MSA) and \RL{المكان رهيب , المحل تحقة , مافي كلام} (dialects),

while `now' is \RL{الان} (MSA) and \RL{هلا , لحين , دلوقت , هسه , هلحزة} (dialects).

\item There are differences in the conjugation of verbs, even though the root is retained. For example, conjugation of the root in MSA is \RL{ل, ع, ب} , while in dialects it is \RL{ لعِبَ}. 

This can be seen in `he plays', which is \RL{هو يلعب} (MSA) and \RL{ يلعب , بيلعب , بلعب } (dialects).

\end{itemize}

The World Arabic Language Dialects map\footnote{\url{https://www.importanceoflanguages.com/arabic-dialects/}} indicates 21 Arabic dialects and shows the different regions of the world in which they are spoken. This can give us hints about how dialects are related to each other.
Table \ref{tb1} (derived from istizada.com\footnote{\url{https://istizada.com/complete-list-of-arabic-speaking-countries-2014/}}) shows the number of speakers for eight of the most important dialects. As can be seen, Sudanese Arabic is the fifth most widely spoken dialect, with $32$ million speakers. This is why we have concentrated on Sudanese in this work.

Sudanese vocabulary is mostly inspired by MSA, but with important Greek, Turkish and English modifications to the phonology. The morphology of Sudanese words shares many features with MSA, but the method of dialect inflection is more complicated than MSA in some respects \cite{hussien2018comparison}. Table \ref{tb3} illustrates the differences between MSA and Sudanese dialect by means of some examples.
Firstly, MSA \RL {اليوم} (`today') corresponds to Sudanese \RL {الليلة} (literally `night'). Even though the word is derived from `night' in MSA,

it nevertheless means `day' in Sudanese. Here, the word meaning is completely reversed.

Secondly, `room' in MSA is \RL {الغرفة} while in Sudanese it is \RL {الأوضة}. Many such nouns are expressed differently. Another example is the names of popular foods and beverages, such as `coffee' (MSA \RL {قهوة } and Sudanese \RL {جبنة}).
Verbs can also differ; the verb `find' in MSA is \RL {تجد} while in Sudanese it is \RL {تشوف}.
Generally, we can see many differences in vocabulary as well as variations in grammar and means of expression.

Following a thorough study of such dialect differences, we have created two datasets based on social media posts, built a CNN-based model for sentiment analyis, and applied it to the datasets.
Below are the main contributions of this work:
 \begin{itemize}
 \item 	We create a public 2-class Sudanese sentiment dataset SudSenti2 from Facebook and YouTube\footnote{\url{https://github.com/mustafa20999/Sudanese-Arabic-Sentiment-Datasets}}.

 \item We build a public 3-class Sudanese sentiment dataset SudSenti3 from Twitter\footnotemark[5].

 \item 	We design a Sudanese stop-word list, and use it for text normalization in the preprocessing phase.

 \item 	We propose a Sentiment Convolutional Model (SCM)  which
 is a five-layer Convolutional Neural Network incorporating our Mean Max Average (MMA) pooling layer.
 
 \item We compare SCM with other machine learning and deep learning methods, and show that it gives a high classification performance.

 \item 	We also compare the proposed MMA pooling layer to the standard pooling layer used in other works, and show that it gives the best performance.

 \item Finally, we apply SCM to datasets in both MSA and Saudi dialect; the proposed approach once again shows a high performance.
\end{itemize}

The paper is organized as follows. Section 2 reviews previous work on sentiment analysis for Arabic. Section 3 describes the creation of the SudSenti2 and SudSenti3 datasets. Section 4 outlines the proposed model architecture. Section 5 presents our experiments, including preprocessing steps, experimental settings, baselines, results and discussion. Finally, Section 6 draws conclusions and suggests future work.

\begin{figure*}[ht]
		\centering
		\includegraphics[width=.99\linewidth]{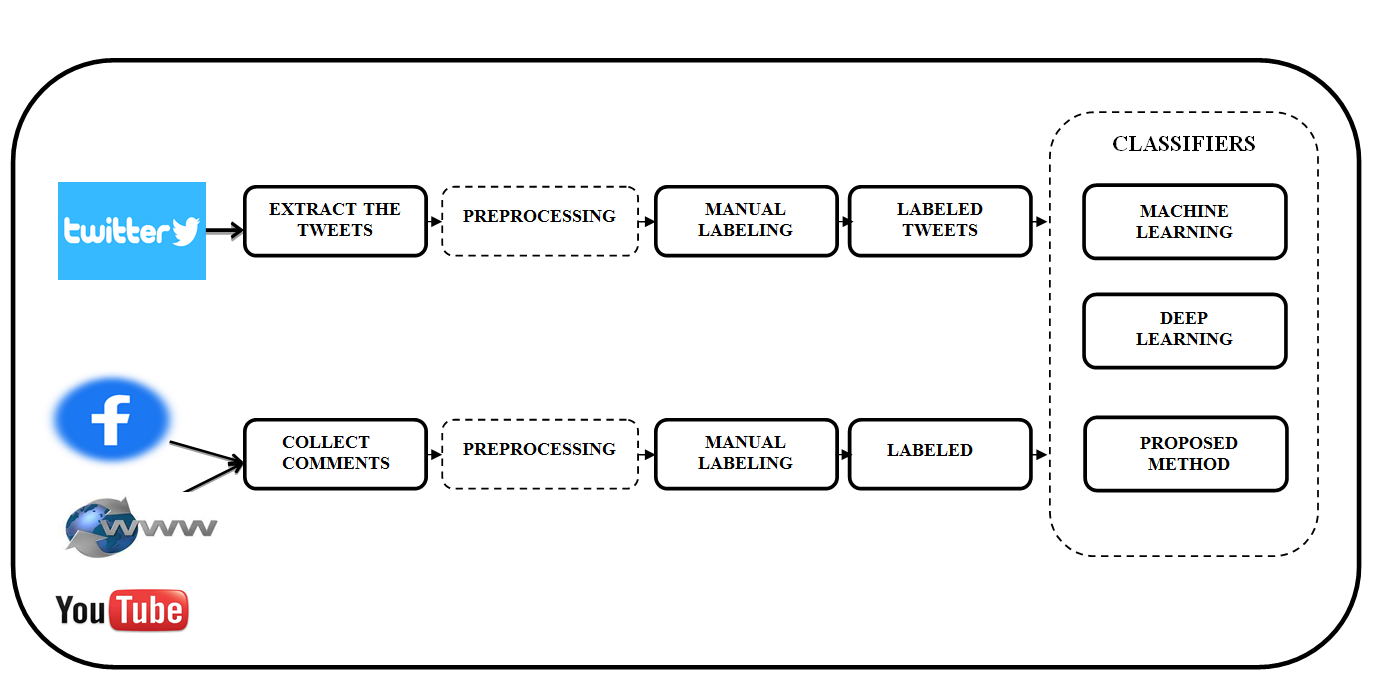}
		\caption{Overall description of work.}
		\label{fig:a1.PNG}
	\end{figure*}

\section{Related work}

As we have mentioned, the Arabic language is very widespread in the world and is spoken in many dialects. Some of these have already been the subject of sentiment analysis research (Table \ref{tb4}).
Here, we discuss which dialects have been studied, what datasets were used, and what sentiment analysis techniques were adopted. For a recent review of Arabic sentiment analysis, please also refer to \cite{alharbi2021deep}.

\cite{tabii2018big} used Naïve Bayes (NB), Maximum Entropy (ME), and Support Vector Machines (SVM) on two datasets,  the Moroccan Sentiment Analysis Corpus (MSAC) \cite{al2018arabic}, comprising tweets from  Twitter and comments from  Facebook and YouTube. SVM was the best single classifier, measured by accuracy (82.5\%). The best
ensemble classifier combined SVM, NB and ME classifiers with majority voting (83.45\%).

\cite{lulu2018automatic} utilized Long-Short Term Memory (LSTM) \cite{hochreiter1997long}\cite{gers1999continual}, Convolutional Neural Networks (CNN), Bidirectional LSTM (BLSTM) and Convolutional LSTM (CLSTM) on three Arabic datasets, Arabic Online Commentary (AOC)\cite{zaidan2011arabic}, Egyptian (EGP), Gulf including Iraqi (GLF), and Levantine (LEV). Results show that LSTM attained the highest accuracy (71.4\%) followed by CLSTM (71.1\%) and BLSTM (70.9\%). It can be observed that the CNN model suffered from overfitting problems as shown by the difference between the cross-validation and test results. 

\cite{al2018sentiment} applied NB, SVM, Decision Trees (DT), and K-Nearest Neighbor (KNN) algorithms on four Arabic datasets, Opinion Corpus for Arabic (OCA) \cite{rushdi2011oca}, Modern Standard Arabic (MSA), Crawler tweets 2014 datasets \cite{abdulla2013arabic}, and the Large-scale Arabic Sentiment Analysis Dataset (LABR)  \cite{aly2013labr}. The aim was to determine the emotions of the Arabic text, using methods based on bigrams and voting combinations. Accuracy was 93.4\%, better than the individual classifiers.
 
\cite{al2019sentiment} used Logistic Regression (LR) with a Term Frequency -- Inverted Document Frequency (TF*IDF) weighting model for feature extraction, on Arabic Services Reviews in Lebanon (ASRL) collected  from Google reviews and the Zomato website in the Lebanon dialect.
For positive classifications, P = 0.88 and R = 1.00, and for negative, P = 0.80, and R = 0.80. Thus the positive result is better than the negative.

\cite{abdelli2019sentiment} utilized SVM and LSTM on both Modern Arabic and the Algerian dialect \cite{moudjari2020algerian}. The results for SVM and LSTM on the Algerian dataset were 86\% and 81\% respectively.

\cite{alwehaibi2018comparison} applied an LSTM Recurrent Neural Network (RNN) (LSTM-RNN) on the AraSenTi dataset which comprises tweets written in MSA and Saudi dialect, manually annotated for Sentiment. Arabic word embeddings used Word2Vec, GloVe and Fasttext \cite{mikolov2013efficient}. The LSTM-RNN model achieved 93.5\% accuracy. 
\cite{alahmary2021semiautomatic} built another Saudi corpus, this time with three classes, produced using a semi-automatic annotation method starting with Naıve Bayes followed by hand correction. They then applied SVM, LSTM, Bi-LSTM, and CNN classifiers. The highest performance was CNN (86.54\%).

\cite{jerbi2019sentiment} used RNN, LSTM, Bi-LSTM, and Deep-LSTM \cite{yilmaz2019deep} on the Tunisian Sentiment Analysis Corpus (TSAC). Deep-LSTM had the highest accuracy (90.00\%).

\cite{mulki2019empirical} experimented with the effect of Named Entity Recognition (NER), using SVM, and NB on four Arabic datasets, Jordanian Egyptian Gulf (JEG), Tunisian Arabic Corpus (TAC), Tunisian Election Corpus (TEC) \cite{sayadi2016tunisian}, and Tunisian Sentiment Analysis Corpus (TSAC) \cite{mdhaffar2017sentiment}. The highest accuracy recorded was on TSAC (82.8\%).

\cite{mhamed2021improving} presented a comprehensive Arabic preprocessing approach, then designed two architectures, MC1 and MC2. On the difficult ASTD dataset \cite{nabil2015astd}, for the 4-class task, accuracy was 73.17\%, on 3-class it was 78.62\%, and on 2-class it was 90.06\%. On the large 2-class ATDFS dataset \cite{alharbi2019}, their model worked effectively, with performance up to 92.96\%.

\cite{al2020meter} use an RNN-based approach on a dataset of Arabic poems (55,440 verses and 14 meters). A five-layer Bidirectional Gated Recurrent Unit (BiGRU) gave the best performance (94.32\%).

\cite{addi2020sampling} applied SVM, NB, and Random Forest (RF) with two techniques, under-sampling and over-sampling. They used the Hotel Arabic Reviews Dataset (HARD) -- Imbalanced \cite{elnagar2018hotel}. RF with Synthetic Minority Oversampling (SMO) gave the best accuracy (96.00\%).

Finally, the task of Arabic preprocessing has received attention in previous work \cite{saad2010impact}\cite{sallam2016improving}\cite{hegazi2021preprocessing} including an approach presented by the authors \cite{mhamed2021improving}.
Here we create two Sudanese Arabic sentiment datasets, one 2-Class and one 3-Class. After detailed preprocessing, we apply the proposed classifier SCM+MMA and compare its performance to ML and NN classifiers.
\clearpage

\section{Dataset Creation}\label{datasets}
In this work, two new datasets for Sudanese are proposed. The 2-Class Sudanese Sentiment Dataset (SudSenti2) was created from Facebook and YouTube. The 3-Class Sudanese Sentiment Dataset (SudSenti3) was created from Twitter posts. Table \ref{tb7} shows summary information for the two datasets.

\subsection{SudSenti2 Dataset}
The following steps were carried out:
\begin{enumerate}
\item Texts were collected from Facebook\footnote{\url{https://www.facebook.com/SudanLovers/}} and YouTube\footnote{\url{https://www.youtube.com/watch?v=h5tBHZZ4UCY&t=1s}}.
\item All texts matching queries such as the following\footnote{`Sudan is a bountiful country', `Diversity of different cultures in Sudan'.} were downloaded, using the Orange Data Mining software\footnote{\url{https://orangedatamining.com/}}: 

\RL{السودان بلد غني} ,
\RL{تنوع الثقافات المختلفة في السودان}
This resulted in 4,544 matching posts.

\item Three judges were chosen to classify the posts. All were university teachers who were native speakers of Sudanese Arabic. All judges judged all posts.
\item Posts which were not considered Sudanese by at least two of the three judges were deleted.

\item Each post was then classified as Negative, Positive or Neutral. (Neutral posts were subsequently deleted.)
A text is considered positive if it contains joyful, happy, or amusing vocabulary, or if there is a positive emoji, or if there is more than one emotion but the positive feeling is dominant. A text is
negative if it contains negative, disappointed, sad, or disturbing vocabulary, or if there is a negative emoji, or if there is more than one feeling and the negative emotion is dominant. Finally, a text is considered neutral if it not clearly positive or negative.
\item Judges worked independently. If at least two of the three judges classified a post as negative it was judged Negative sentiment, and similarly for Positive sentiment. Neutral posts were deleted from the collection, resulting in a 2-class dataset.
\item By following the above procedure, 4,000 posts were selected from the original 4,544. The final SudSenti2 dataset contains 2,027 positive posts and 1,973 negative posts.

\end{enumerate}

\subsection{SudSenti3 Dataset}
The following steps were carried out to produce the 3-class dataset:
\begin{enumerate}
\item All Twitter messages matching one of the search strings mentioned above were downloaded using the Twitter API. This resulted in 8,021 posts.
\item The same three judges classified the posts as for SudSenti2.
\item Posts not considered Sudanese by at least two of the three judges were deleted.
\item Each tweet was classified as Positive, Negative or Neutral. SudSenti3 is thus a 3-class dataset.
\item Judges worked independently. Posts classified positive by at least two judges were considered Positive, and the same for Negative and Neutral. Posts where there was no majority judgement were eliminated.
\item By following the above procedure, 7,109 tweets were selected from the original 8,021. The resulting SudSenti3 dataset contains 2,523 positive posts and 2,639 negative posts.
\end{enumerate}

\begin{table}
\begin{center}

\caption{Examples from the Sudanese stopword list.}\label{tb6}
	\begin{tabular}{ cc } 
		\toprule
		 \midrule
		\RL{هسع, هسي, هسه, اسي, حسع, حسة} & now\\ \hline
		\RL{ هدا } & this\\ \hline
	 \RL{ إنحنا } & we\\ \hline\ 
		\RL{  كلو } & everything\\ \hline
\RL{  هنديلكم, هنديلكن } & them (men), them (women)\\ \hline
\RL{ ديلكم, دة , داك , دييكه }& them (medium group), this (one person close),\\
& that (one male person far), that (one female person far) \\
\hline
\RL{ هنديك, هنداك } & he, she (faraway whom you know)\\ \hline
		 \RL{ وين } & where\\
		 \hline
  \bottomrule
	\end{tabular}
\end{center}
\end{table}

\section{Proposed Approach}
\subsection{Text Preprocessing}
\begin{enumerate}
\item The following information is removed from each post: URL links, account name, description, time of data creation, followers, profile image URL, location, screen name, favourites, friends.
\item @date and @time symbols are removed.
\item Punctuation marks and diacritics are removed \cite{hegazi2021preprocessing}.
\item Strip elongation is carried out, changing
`\RL { نـــــــــــــعوســـــــــــــه}' into `\RL { نعوسة }'\footnote{Local process of cooking food.}.
\item Heh normalization is carried out, e.g.`\RL{ة}' becomes `\RL{ه}'. 
Similarly for Yeh normalization, `\RL{ي}' to `\RL{ى}',  

Caf normalization, `\RL{ك}' to `\RL{كـــ}' , Hamza normalization, `\RL{ئ}' or `\RL{ؤ}' to `\RL{ء}', 

and Alef normalization, `\RL{آ}' or `\RL{أ}'or `\RL{إ}' to `\RL{ا}'
\cite{sallam2016improving}.

\item Redundant letters like `\RL{ عااااااجل}' are removed.
\item Numbers and non-Arabic letters are removed.
\item Stopwords are removed.
\end{enumerate}

A stopword list has been produced containing MSA and colloquial Arabic stopwords used in Sudan. The list contains 269 words and 2,095 characters (Table \ref{tb6}\footnote{English translations are indicative only, and are assuming Sudanese interpretations.}).

\subsection{Text Encoding}
\textbf{Input layer:}
In order to start, let us assume that the input layer receives text data as $X(x 1, x 2,..., x n)$, where $x 1, x 2,..., x n$ is the number of words with the dimension of each input term $m$. Each word vector would then be defined as the dimensional space of $R^m$. Therefore, $\mathbb{R}^{m\times n}$ will be the input text dimension space.

\textbf{Word embedding layer:}
Let us say the vocabulary size is $d$ for a text representation in order to carry out word embedding. Thus, it will represent the dimensional term embedding matrix as $A^{m\times d}$. The input text $X(x I )$, where $I = 1, 2, 3,..., n$, $X$ $\epsilon$ $\mathbb{R}^{m\times n}$, is now moved from the input layer to the embedding layer to produce the term embedding vector for the text. For the dialects we use TF-IDF, and for MSA we implemented the AraVec \cite{soliman2017aravec} word embedding pre-trained by Word2Vec \cite{mikolov2013distributed} on Twitter text.

The representation of the input text $X(x_1, x_2,..., x_n )$ $\epsilon$ $\mathbb{R}^{m\times n}$  as numerical word vectors is then fed into the model. $x_1, x_2,...,x_n$ is the number of word vectors with each dimension space $R^m$ in the embedding vocabulary.

\subsection{Mean Max Average Pooling}
In CNNs, the pooling function is essential for extracting the specific features from the feature map. The aim of pooling is to determine the output of  $Y_k$, the pooling $P_k$ for $k =1,..., K$, where the set of activations in $P_k$ is represented as $c_1,...,c_{\vert P_k\vert}$ and $\vert P_k\vert$ denotes the number of activations. By collecting the outputs of all the pooling regions, the pooling feature map E= $e_1,...,e_k$ is obtained. We will start with a quick overview of standard pooling strategies:

Max pooling \cite{ranzato2007sparse}: This takes the biggest activation in the pooling region:
\begin{equation}
Max_K = \max_{i \epsilon P_k c_i}, \quad \quad k=1,..., K.  %
\end{equation}

Max pooling is ideal for extracting local characteristics from a feature map, such as edges, lines, and textures.

Average pooling \cite{lecun1989handwritten}: This calculates the mean value of activities in the pooling region:
\begin{equation}
    Avg_K  = \frac{1} {\vert P_k\vert}\sum_{i \epsilon P_k }c_i, \quad  \quad k=1,..., K
\end{equation}

By smoothing the pooling region in this way, it is possible to extract global characteristics.

Min pooling \cite{socher2011dynamic}: Calculates the minimum value of activities in the pooling region:

\begin{equation}
    Min_K= \min_{i \epsilon P_k c_i}, \quad  \quad k=1,..., K
\end{equation}

\begin{figure*}[ht]
		\centering
		\includegraphics[width=.99\linewidth]{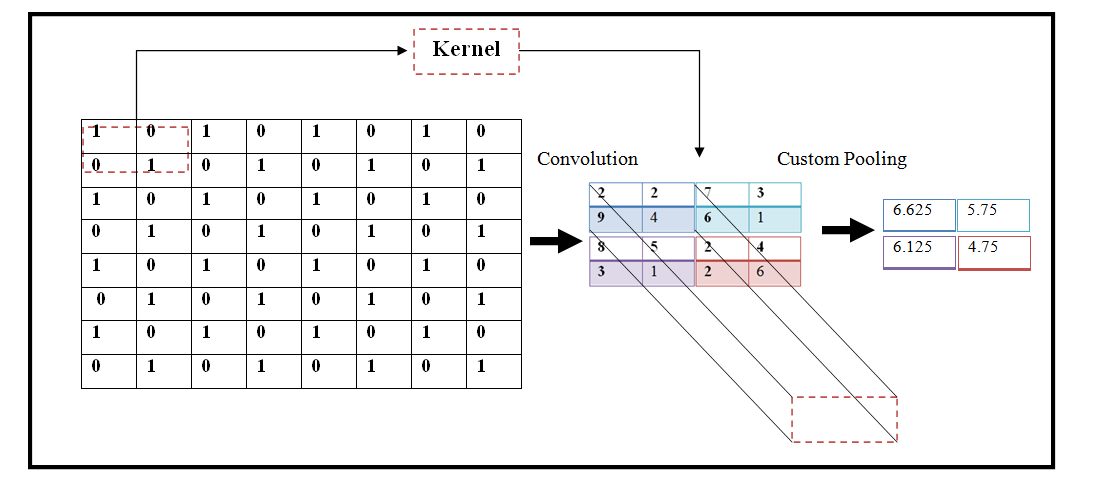} 
		\caption{Mean Max Average (MMA).}
		\label{fig:a3.PNG}
	\end{figure*}

Our proposed Mean Max Average (MMA, Figure \ref{fig:a3.PNG}) pooling calculates the mean of the max value and the average value:

\begin{equation}
    MMA_K=\frac{(\max_{i \epsilon P_k c_i} ) +\left(\frac{1}{\vert P_k\vert}\sum_{i \epsilon P_k }c_i  \right)}{2}, \\
    \nonumber  \quad k=1,..., K
\end{equation}

MMA aims to combine the advantages of Max pooling and Average pooling.

\begin{figure*}[ht]
		\centering
		\includegraphics[width=.99\linewidth]{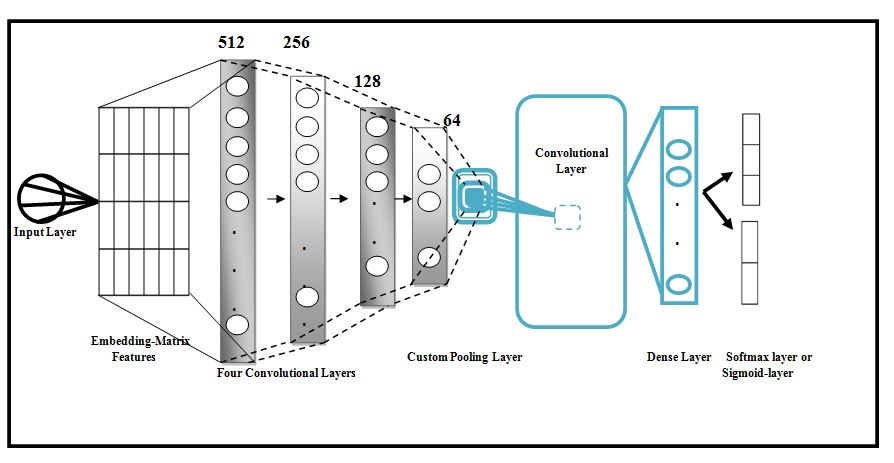} 
		\caption{SCM Model Architecture.}
		\label{fig:a2.JPG}
	\end{figure*}

\subsection{Proposed Approach}
SCM (Figure \ref{fig:a2.JPG}) consists of an Embedding layer containing max-features = num-unique-word, embedding-size [128, 300] with max-len [150, 80, 50]. After that, there are four Convolutional neural network layers with filters respectively of [512, 256, 128, 64]. kernel-size = 3, padding = `valid', activation = `ReLU', and strides = 1. These are followed by the proposed Mean Max Average (MMA) pooling function, 1D pool size = 2, then Dense = 32, and activation=`ReLU', then Dropout 0.5, then batch normalization and another Dropout 0.5, then Flatten and finally a softmax layer. This is fully connected to predict the sentiment between three classes (Positive, Negative, Neutral) or two classes (Positive, Negative).

\begin{table*}
\centering
		\caption{Datasets for our experiments.}\label{tb7}
		\begin{tabular}{lllllll}
			\toprule
			Datasets    & positive tweets & negative tweets  & neutral tweets & Total \\
			\midrule
			SudSenti2  (2C) & 2,027 & 1,973  & - & 4,000 \\
			SudSenti3 (3C) & 2,523 & 2,639  & 1,947 & 7,109 \\
			SSD   (2C) & 2,436 & 1,816  & - & 4,252 \\
			HARD  (2C) & 5,857 & 6,353  & - & 12,210  \\
		
			\bottomrule
		\end{tabular}
	\end{table*}

\begin{table}
	\caption{Experimental settings.}\label{tb8}
	\centering
	\begin{tabular}{llll}
		\toprule
		Setting  & Value(s)\\
		\midrule
		embedding size & \{\textbf{100}, \textbf{128}, 200, \textbf{300}\}\\
        pooling & \{\textbf{2}, \textbf{4}, \textbf{{6}}, 8, 16\}\\
        batch-size & \{\textbf{32},\textbf{64}, \textbf{128}, \textbf{164}, \textbf{200}, 400\}\\
        kernel-size & \{\textbf{3}, \textbf{5}, 7, 10\}\\
        number-classes & \{\textbf{2}, \textbf{3}, {4}, 5, 10\}\\
        epoch & \{\textbf{5}, \textbf{10}, 20, \textbf{50}, \textbf{100}, 200\}\\
        optimizer & Adam\\
        learning rate & \{0.01, \textbf{0.001}, 0.0001\}\\
		\bottomrule
	\end{tabular}
\end{table}

\section{Experiments}
Our experiments include four aspects (Figure \ref{fig:a1.PNG}):
\begin{enumerate}
\item Preprocessing the datasets and checking the steps.
\item Utilizing existing machine learning and deep learning methods to verify performance. 
\item Applying the proposed method.
\item Analyzing results.
\end{enumerate}

\subsection{Datasets}
For sentiment classification on Sudanese Arabic text, machine learning and deep learning models are trained using the new
SudSenti2 and SudSenti3 datasets introduced in Section \ref{datasets}.

SudSenti2 consists of two classes, 2,027 positive tweets and 1,973 negative tweets. SudSenti3 consists of three classes, 2,523 positive tweets, 2,639 negative tweets, and 1,947 neutral tweets.

For the Saudi dialect we use the Saudi Sentiment Dataset (SSD) \cite{alyami2020application}\footnote{\url{https://www.kaggle.com/snalyami3/arabic-sentiment-analysis-dataset-ss2030-dataset}}.

SSD consists of two classes, 2,436 positive tweets and 1,816 negative tweets.

For sentiment classification in MSA, the models are trained using the Hotel Arabic reviews dataset (HARD)\footnote{\url{https://github.com/elnagara/HARD-Arabic-Dataset} produced by \cite{elnagar2018hotel}}. It is a rich dataset, with more than 370,000 reviews expressed in MSA. Here we utilized two classes, 5,857 positive tweets and 6,353 negative tweets.

Table~\ref{tb7} shows the details of the datasets.

\subsection{Experimental Settings}
We used machine learning algorithms and deep learning models for training with all the Arabic sentiment datasets for 2-way and 3-way classification.
The machine learning algorithms were Naive Bayes (NB) \cite{mohammad2019arabic}, Logistic Regression (LR) \cite{al2016building}, Support vector machines (SVMs) \cite{mohammad2016arabic} and Random forest (RF) \cite{ravi2017novel}.

 Deep learning models include RNN \cite{britz2015recurrent}, CNN \cite{zhang2015sensitivity}, CNN-LSTM \cite{smaili2019arabic}	\footnote{\url{https://www.kaggle.com/monsterspy/conv-lstm-sentiment-analysis-keras-acc-0-96}} and the proposed method. 
 For the SudSenti2 and SudSenti3 datasets, we split the data into 80\% training, 10\% validation and 10\% testing.

For the SSD and HARD datasets, we applied the proposed approach and existing deep learning models. The settings for the experiments are shown in Table~\ref{tb8}. We used our own tuning and hyperparameter values and chose the TensorFlow framework for the implementation.

\begin{table*}
	\caption{Experiment 1: Accuracy of ML and NN sentiment classifiers on 2-class datasets. SudSenti2 is a new 2-class dataset for Sudanese, created from Facebook and YouTube (Section \ref{datasets}). SCM+MMA is the proposed model.}\label{tb9}
	\centering
	\begin{tabular}{lllll}
		\toprule
		Models    & Accuracy (\%) &  &    \\
		\toprule
		    &  SudSenti2 Dataset (2C) & SSD  Dataset (2C) & HARD Dataset (2C) \\
		\midrule
			LR & 86.04 & - &  - \\
		RF & 87.12  & - &  -  \\
		NB & 81.45  & - &  -  \\
		SVM & 86.23 & - &  - \\
		RNN & 80.75  & 74.03 & 62.86  \\
		CNN & 87.75  & 82.49 & 87.06   \\
		CNN-LSTM & 89.00 & 83.55 & 85.22   \\
		SCM+MMA & \textbf{92.25} & \textbf{84.02} & \textbf{88.37}  \\
		\bottomrule
	\end{tabular}
\end{table*}

\begin{table*}
	\caption{Experiment 2: Accuracy of NN sentiment classifiers on the SudSenti3 3-class dataset, created from Sudanese Twitter posts (Section \ref{datasets}). SCM+MMA is the proposed model.}\label{tb10}
	\centering
	\begin{tabular}{ll}
		\toprule
		Models    & Accuracy (\%)     \\
		\toprule
		    &   SudSenti3 Dataset (3C)  \\
		\midrule
		LR & 79.37   \\
		RF & 78.24    \\
		NB & 74.19    \\
		SVM & 79.29    \\
		RNN & 77.07   \\
		CNN & 83.61    \\
		CNN-LSTM & 81.01   \\
		SCM+MMA & \textbf{85.23}    \\
		\bottomrule
	\end{tabular}
\end{table*}

\begin{table*}
	\caption{Experiment 3: Accuracy of the SCM model with different pooling layers. The task is 2-class sentiment classification, applied to the SudSenti2, SSD, and HARD datasets. MMA is the proposed pooling layer.}\label{tb11}
	\centering
	\begin{tabular}{lllll }
		\toprule
		Models    & Accuracy (\%) &  &    \\
		\toprule
		    &  SudSenti2 Dataset (2C) & SSD  Dataset (2C) & HARD Dataset (2C) \\
		\midrule
		SCM+Max & 90.62  & 84.72 & 89.27  \\
		SCM+Avg & 91.75  & 84.02 & 87.80   \\
		SCM+Min & 90.00 & 83.78 & 88.70   \\
		SCM+MMA & \textbf{92.75} & \textbf{85.55} & \textbf{90.01}  \\
		\bottomrule
	\end{tabular}
\end{table*}
\begin{table*}
	\caption{Experiment 3: Accuracy of the SCM model with different pooling layers. The task is 3-class sentiment classification, applied to the SudSenti3 dataset. MMA is the proposed pooling layer.}\label{tb12}
		\centering
	\begin{tabular}{ll}
		\toprule
		Models    & Accuracy (\%)     \\
		\toprule
		    &   SudSenti3 Dataset (3C)  \\
		\midrule
		SCM+Max & 84.11   \\
		SCM+Avg & 83.26    \\
		SCM+Min & 82.70   \\
		SCM+MMA & \textbf{84.39}    \\
		\bottomrule
	\end{tabular}
\end{table*}

\subsection{Experiment 1: Two-way Sentiment Classification.}
The aim was to evaluate the proposed SCM+MMA model in 2-way sentiment classification, working with the SudSenti2, SSD and HARD datasets. SudSenti2 was introduced in Section \ref{datasets}. As baselines there are four ML models (LR, RF, NB, SVM) and three NN models (RNN, CNN, CNN-LSTM).

The configuration of SCM+MMA is shown in Table \ref{tb8}. Ten-fold cross-validation was used for all models and the average performance reported. The results are shown in Table \ref{tb9}.

On SudSenti2 (Sudanese dialect), the best model is SCM+MMA (accuracy 92.25\%). The best ML baseline was RF (87.12\%) and the best NN baseline was CNN-LSTM (89.00\%).

On the SSD dataset (Saudi dialect), the best model is SCM+MMA (84.02\%) and the best baseline is CNN-LSTM (83.55\%). Finally, on the HARD dataset (MSA) the best model is again SCM+MMA (88.37\%) as against the best baseline, CNN (87.06\%).

In summary, the experiment showed that the proposed model performed well on two-class datasets.

\subsection{Experiment 2: Three-way Sentiment Classification}
The aim was to evaluate SCM+MMA once again, this time on the new 3-way Sudanese dataset, SudSenti3 (Section \ref{datasets}). 3-way classification is known to be a harder task than 2-way, particularly as the Neutral class can contain examples with both positive and negative aspects, a factor which may confuse the model.
As baselines there are three NN models (RNN, CNN, CNN-LSTM). The configuration of SCM+MMA was the same as in Experiment 1 (Table \ref{tb8}) except that there were three outputs, not two. Once again, ten-fold cross-validation was used for all models. The results are shown in Table \ref{tb10}.

The best performing model is SCM+MMA (85.23\%). The best ML baseline is LR (79.37\%) and the best NN baseline is CNN (83.61\%).

\subsection{Experiment 3: Evaluation of MMA Pooling}
A key part of the proposed SCM+MMA model is the MMA pooling layer. The aim of this experiment, therefore, was to compare MMA with three commonly-used pooling layers, Max, Avg and Min.
Firstly, in 2-way classification, the performance of SCM+Max, SCM+Avg and SCM+Min was compared with SCM+MMA on SudSenti2, SSD and HARD (compare with Experiment 1). Fifteen-fold cross-validation was used and the average performance reported. Results are shown in Table \ref{tb11}. SCM+MMA is the best performing model on all three datasets (SudSenti2 92.75\%, SSD 85.55\%, HARD 90.01\%). The best baseline pooling layer varies between Avg and Max by dataset (SudSenti2: Avg 91.75\%; SSD: Max 84.72\%; HARD: Max 89.27).

Secondly, in 3-way classification, the performance of SCM+Max, SCM+Avg and SCM+Min was compared with SCM+MMA on SudSenti3 (compare with Experiment 2). 
Fifteen-fold cross-validation was again used. Results are shown in Table \ref{tb12}.
Once again, SCM+MMA is the best performing model (84.39\%) with the best baseline being Max (84.11\%).

In conclusion, the MMA pooling layer performs well compared to Max, Avg and Min.

\subsection{Accuracy during training}
\begin{figure*}[ht]
		\centering
		\includegraphics[width=.99\linewidth]{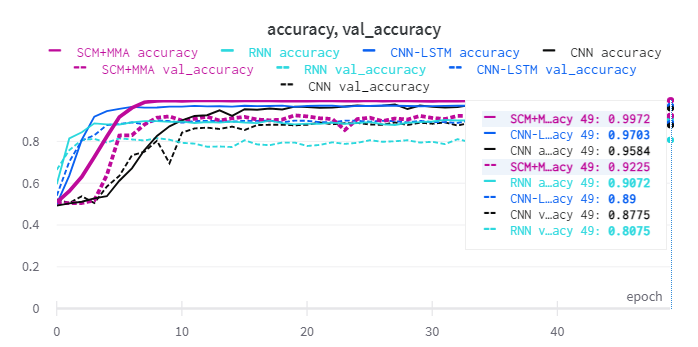}
		\caption{Accuracy and validation accuracy with the SudSenti2 dataset.}
		\label{fig:a4.png}
	\end{figure*}

\begin{figure*}[ht]
		\centering
		\includegraphics[width=.99\linewidth]{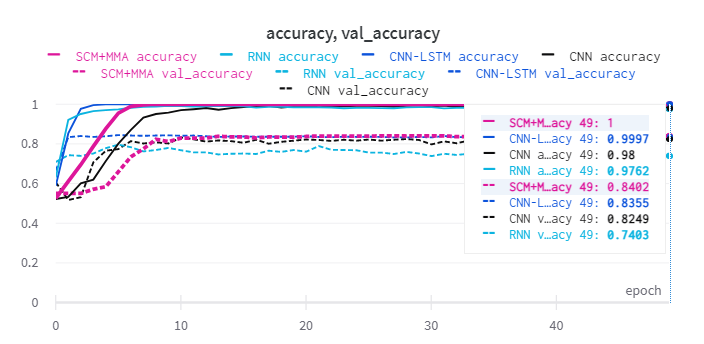}
		\caption{Accuracy and validation accuracy with the SSD dataset.}
		\label{fig:a5.png}
	\end{figure*}

\begin{figure*}[ht]
		\centering
		\includegraphics[width=.99\linewidth]{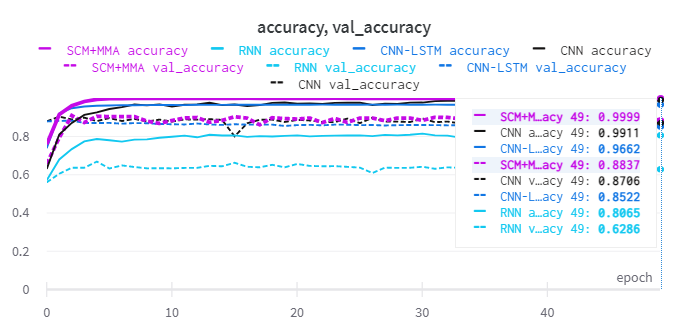}

		\caption{Accuracy and validation accuracy with the HARD dataset.}
		\label{fig:a6.PNG}
	\end{figure*}

Figure \ref{fig:a4.png} shows the accuracy and validation accuracy of the NN baseline models and the proposed method with the SudSenti2 dataset. After 50 epochs, the SCM+MMA model shows the highest performance, reaching 92.25\%. Figure \ref{fig:a5.png} shows the same information for the SSD dataset (SCM+MMA reaches 84.02\%) while Figure \ref{fig:a6.PNG} is for the HARD dataset (SCM+MMA reaches 88.37\%).

\begin{figure*}[ht]
		\centering
		\includegraphics[width=.99\linewidth]{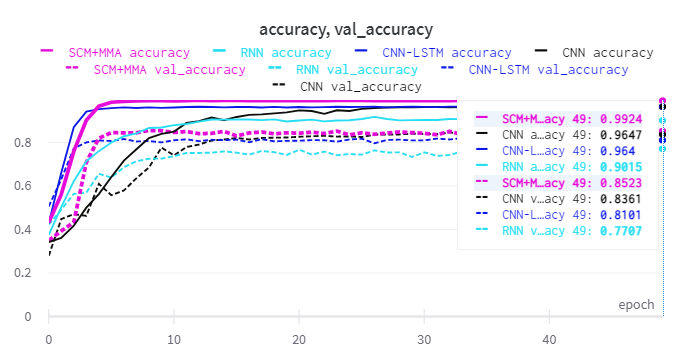}
		\caption{Accuracy and validation accuracy with the SudSenti3 dataset.}
		\label{fig:a7.png}
	\end{figure*}

Figure \ref{fig:a7.png} shows the accuracy and validation accuracy for the NN models and the proposed model with the SudSenti3 dataset. After 50 epochs, SCM+MMA reaches 85.23\%.

\begin{figure*}[ht]
		\centering
		\includegraphics[width=.99\linewidth]{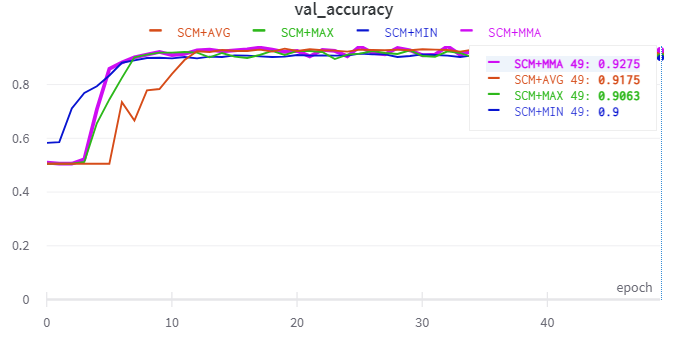}
		\caption{Validation accuracy with the SudSenti2 dataset.}
		\label{fig:a8.png}
	\end{figure*}

\begin{figure*}[ht]
		\centering
		\includegraphics[width=.99\linewidth]{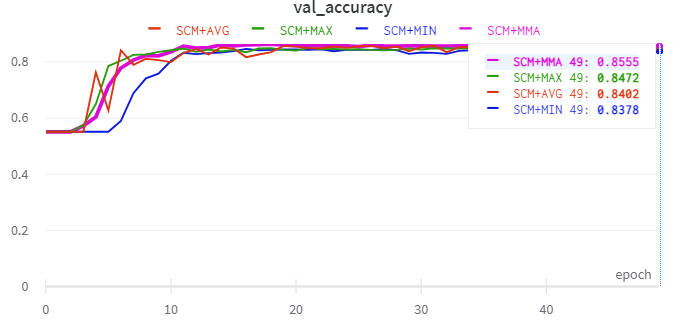}
		\caption{Validation accuracy with the SSD dataset.}
		\label{fig:a9.png}
	\end{figure*}

\begin{figure*}[ht]
		\centering
		\includegraphics[width=.99\linewidth]{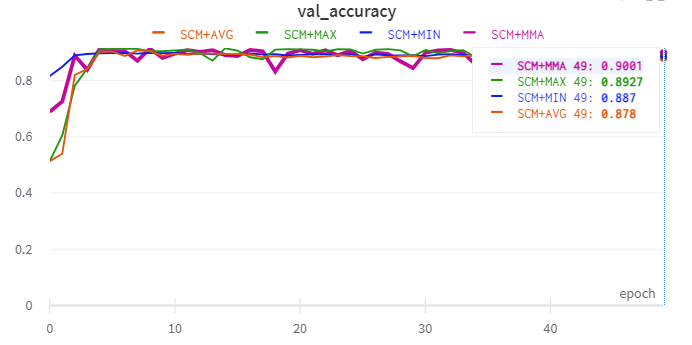}
		\caption{Validation accuracy with the HARD dataset.}
		\label{fig:a10.png}
	\end{figure*}

\begin{figure*}[ht]
		\centering
		\includegraphics[width=.99\linewidth]{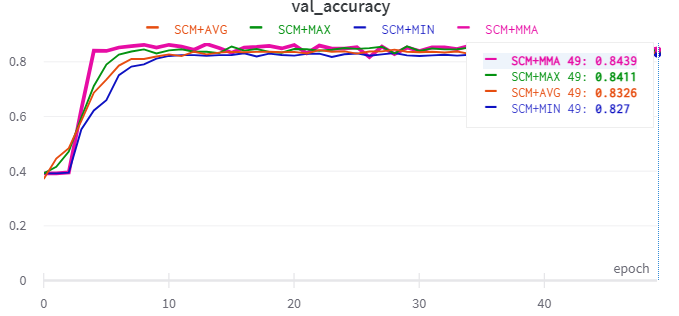}
		\caption{Validation accuracy with the SudSenti3 dataset.}
		\label{fig:a11.png}
	\end{figure*}

For 2-way classification, Figures \ref{fig:a8.png}, \ref{fig:a9.png} and \ref{fig:a10.png} show the validation accuracy during training for the SudSenti2, SSD and Hard datasets.
Finally, for 3-way classification, Figure \ref{fig:a11.png} shows the validation accuracy during training for SudSenti3.
We note that the proposed method was stable over epochs for training and validation with different datasets.

\section{Conclusion and Future Work}
In this paper, we first presented two new sentiment datasets for the Sudanese dialect of Arabic. SudSenti2 was collected from Facebook and YouTube, while SudSenti3 was based on Twitter tweets. Following a discussion of Arabic pre-processing methods appropriate to sentiment classification, we proposed a new model for this task, SCM+MMA. This includes four convolutional layers plus MMA, our proposed pooling layer.
In 2-way sentiment classification using the SudSenti2 (Sudanese), SSD (Saudi) and HARD (MSA) datasets, SCM+MMA gave the best performance relative to ML and NN baselines. In 3-way classification using SudSenti3, SCM+MMA was also superior to the baselines. Finally, the proposed MMA pooling was compared to Max, Avg and Min baselines and shown to perform better than them in both 2-way and 3-way classification.

In future work, we plan to use an attention mechanism as part of a more complex deep learning method, to extract features from a huge corpus covering all Arabic sentiment dialects. 

\section{Data Availability}

The SudSenti2 and SudSenti3 datasets are publicly available\footnote{\url{https://github.com/mustafa20999/Sudanese-Arabic-Sentiment-Datasets}}.

\section{Acknowledgments}

The research for this paper is supported by the Science Support Agency (Grant no. 12345) and the Projects Fund (Grant no. 23456). Many thanks to George Kour for the ArXiv style: https://github.com/kourgeorge/arxiv-style.

\clearpage

\clearpage

\bibliographystyle{abbrv}

\end{document}